\documentclass[lettersize,journal]{IEEEtran}
\usepackage{amsmath,amsfonts}
\usepackage{algorithmic}
\usepackage{algorithm}
\usepackage{array}
\usepackage[caption=false,font=normalsize,labelfont=sf,textfont=sf]{subfig}
\usepackage{textcomp}
\usepackage{stfloats}
\usepackage{url}
\usepackage{verbatim}
\usepackage{graphicx}
\usepackage{cite}
\usepackage{multirow}
\usepackage[table,xcdraw]{xcolor}
\usepackage[normalem]{ulem}
\useunder{\uline}{\ul}{}

\begin{document}

\title{HODINet: High-Order Discrepant Interaction Network for RGB-D Salient Object Detection}

\author{Kang Yi, Jing Xu, Xiao Jin, Fu Guo and Yan-Feng Wu 

\thanks{Manuscript received 14 June 2022. This work was supported in part by Tianjin Natural Science Foundation, China (22JCQNJC01650, 19JCQNJC00300), and Fundamental Research Funds for the Central Universities of Nankai University (63201192, 63211116). (Kang Yi and Jing Xu contributed equally to this work.) (Corresponding author: Xiao Jin.) 

Yi Kang, Jing Xu, Xiao Jin and Yan-Feng Wu are with the College of Artificial Intelligence, Nankai University, Tianjin 300350, China (e-mail: jinxiao@nankai.edu.cn).

Guo Fu is with Tandon School of Engineering, The New York University, New York 11201, USA.

}
}


\maketitle

\begin{abstract}
RGB-D salient object detection (SOD) aims to detect the prominent regions by jointly modeling RGB and depth information. Most RGB-D SOD methods apply the same type of backbones and fusion modules to identically learn the multimodality and multistage features. However, these features contribute differently to the final saliency results, which raises two issues: 1) how to model discrepant characteristics of RGB images and depth maps; 2) how to fuse these cross-modality features in different stages. In this paper, we propose a high-order discrepant interaction network (HODINet) for RGB-D SOD. Concretely, we first employ transformer-based and CNN-based architectures as backbones to encode RGB and depth features, respectively. Then, the high-order representations are delicately extracted and embedded into spatial and channel attentions for cross-modality feature fusion in different stages. Specifically, we design a high-order spatial fusion (HOSF) module and a high-order channel fusion (HOCF) module to fuse features of the first two and the last two stages, respectively. Besides, a cascaded pyramid reconstruction network is adopted to progressively decode the fused features in a top-down pathway. Extensive experiments are conducted on seven widely used datasets to demonstrate the effectiveness of the proposed approach. We achieve competitive performance against 24 state-of-the-art methods under four evaluation metrics.
\end{abstract}

\begin{IEEEkeywords}
RGB-D SOD, cross-modality fusion, high-order attention mechanism, transformer, discrepant interaction.
\end{IEEEkeywords}

\section{Introduction}
\IEEEPARstart{S}{ALIENT} object detection (SOD) refers to locating the most attractive parts of a given scene \cite{Wang2021TPAMI,Cong2019TCSVT}. It plays an important role in the field of computer vision and has been applied to various scenarios. Although the SOD methods have made notable progress and improvement, the performance often degrades when facing complex and indistinguishable cases. With the rapid development of 3D image sensors, people can obtain high-quality depth information easily, which has attracted growing interest among researchers in RGB-D SOD methods. Numerous studies have shown that utilizing RGB-D image pairs for saliency detection can achieve superior performance in some challenging scenes. 

\subsection{Motivation}
The motivation of this work mainly derives from two aspects: On one hand, even though RGB images and depth maps contain completely different information, most of the existing RGB-D SOD methods still employ the same type of backbones to identically extract features from different modalities \cite{DUT_DMRA2019ICCV}. On the other hand, different stages of backbones focus on extracting distinct spatial and channel features, most of the existing RGB-D SOD methods only apply the same kind of fusion modules \cite{liu2021tritransnet,DCF}. Concretely, there still remains a few problems that demand further discussions and improvements in RGB-D SOD.

\begin{figure}[!t]
	\centering
	\includegraphics[scale=0.6]{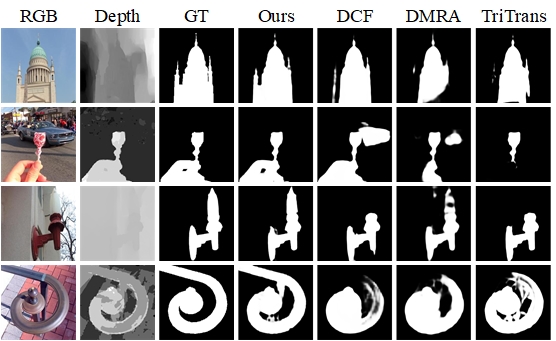}
	\centering
	\caption{The results of our HODINet and some other first-order fusion methods, including DCF \cite{DCF}, DMRA \cite{DUT_DMRA2019ICCV} and TriTransNet \cite{TriTransNet}. Note that TriTransNet \cite{TriTransNet} adopts transformer-based backbones, we still obtain competitive results.}
	\label{fig:examples}
\end{figure}


\subsubsection{how to model discrepant characteristics of RGB images and depth maps}
The RGB images contain local color and texture contents, while the depth maps provide shape and position clues in a global view \cite{Zhang2021MM,Li2020ECCV}. The RGB and depth features with mismatched information will generate suboptimal fusion outcomes, and result in a negative effect on the performance. Most previous methods employ the same type of backbones or siamese networks to extract features identically, which is lack of emphasis on the complementary information between different modalities. One possible solution to this problem is to adopt more specific backbones based on the characteristics of the modalities. For example, local continuity and structure information should be emphasized when encoding RGB images with rich texture features. Since the practical receptive fields are much smaller than the theoretical ones \cite{Luo2016NeurIPS}, the backbone for the depth branch needs to take a stack of convolutional layers to include global position relationship rather than local details.


\subsubsection{how to fuse features from two modalities in different stages}
Spatial and channel attention plays a crucial role in fusion modules and brings significant performance improvements. Nevertheless, due to the discrepancy in resolutions and channels of features maps in different stages, utilizing the same fusion module in all stages cannot guarantee satisfactory results. The neural networks tend to obtain some specific features in the shallow layers, while some abstracted semantic features are learned in the deep layers. 
Different types of features should be discrepantly integrated and fused to achieve better experimental results.
Therefore, it is significant to design suitable fusion modules that take into account the diversity of different stages. 

\subsection{Contribution}

To tackle the above issues, we propose a high-order discrepant interaction network (HODINet) for RGB-D SOD. 
It is a dual-stream asymmetric framework where transformer-based and CNN-based architectures are used as backbones to extract features from RGB images and depth maps.
Moreover, high-order statistical information has been proved to be able to distinguish subtle differences between images in fine-grained image classification \cite{Zheng2019CVPR,Kong2017CVPR}. Inspired by this, to exploit complementary information between different modalities, high-order representations are carefully embedded into spatial and channel attention to fuse the features from RGB images and depth maps. 
Since low-level features contain spatial information and high-level features concentrate on channel dimensions, a high-order spatial fusion (HOSF) module and a high-order channel fusion (HOCF) module are introduced to provide detailed and accurate guidance for feature fusion. Besides, an effective cascaded pyramid reconstruction network (CPRN) is adopted for cross-stage decoding. Extensive experiments are conducted on seven benchmark datasets, and the results indicate that the proposed HODINet achieves competitive performance against 24 RGB-D SOD methods. Fig. \ref{fig:examples} illustrates a comparison of detection results of our HODINet and other conventional fusion methods.

In summary, the main contributions in this paper can be described as follows.

\begin{itemize}
	\item We present a high-order discrepant interaction network (HODINet) for RGB-D SOD, which is a dual-stream asymmetric framework using transformer-based and CNN-based architectures to extract features from RGB images and depth inputs.
	\item To distinguish complementary information between two modalities, we design a high-order spatial fusion (HOSF) module and a high-order channel fusion (HOCF) module, which learn high-order representations in spatial and channel attention.
	\item We also develop a cascaded pyramid reconstruction network (CPRN) to progressively decode the multi-scale fused features in a top-down pathway.
	\item The proposed method is evaluated on seven publicly available datasets under four widely used metrics. Compared with 24 state-of-the-art approaches, HODINet shows comparable performance. 
\end{itemize}

The rest of this paper is organized as follows. In Section II, the related work is presented. Then, the detailed methodology is described in Section III. The experimental settings and result discussion are shown in Section IV. Finally, Section V concludes this paper.

\section{Related Work}
\subsection{RGB-D Salient Object Detection}
Early RGB-D SOD methods usually design handcrafted features and various fusion strategies to integrate RGB and depth inputs. Following this direction, numerous models have been proposed to detect salient objects \cite{peng2014rgbd,song2017depth,feng2016local,ren2015exploiting}. However, they merely regard the depth stream as auxiliary information, which results in unsatisfactory performance.

With the rapid development of deep learning, CNN-based methods have achieved remarkable progress \cite{chen2019multi,CPFP,SIP_D3Net2021TNNLS,A2dele2020CVPR}. Huang et al. \cite{EBFS} explore cross-modality interactions between RGB and depth features using linear and bilinear fusion strategies. Zhang et al. \cite{DRSD} estimate depth maps by the corresponding RGB images and integrate depth features to enhance the saliency detection performance. Huang et al. \cite{JCUF} capture cross-level and cross-modal complementary information by multi-branch feature fusion and selection. Jiang et al. \cite{cmSalGAN} incorporate intra-view and correlation information in RGB-D SOD by a novel cross-modality saliency generative adversarial network. Liu et al. \cite{Liu2021TMM} build an attentive cross-modal fusion network based on residual attention for RGB-D SOD. Zhou et al. \cite{Zhou2021TMM} present a cross-level and cross-scale adaptive fusion network to detect RGB-D saliency. Mao et al. \cite{Mao2021TMM} propose a novel cross-modality fusion and progressive integration network to address saliency prediction on stereoscopic 3D images.


With the recent success in other computer vision tasks \cite{dosovitskiy2020vit}, the transformer-based models have opened up new directions in RGB-D SOD. Liu et al. \cite{liu2021visual} design a pure transformer framework, which utilizes the T2T-ViT to divide images into patches and the RT2T transformation to decode patch tokens to saliency maps. Liu et al. \cite{liu2021tritransnet} introduce a triplet transformer scheme to model the long-range relationship among different stages, which serves as supplementary encoding strategy for feature fusion. Liu et al. \cite{SwinNet} propose SwimNet for RGB-D SOD. They fully explore the advantages of CNN and transformer in modeling local and global features. 
Due to the inherent discrepancy of the two modalities, using the exact same type of backbones to extract features indiscriminately may not obtain satisfactory results. Thus, the proposed HODINet is designed as a dual-stream asymmetric framework using transformer-based and CNN-based architectures to extract features from RGB images and depth maps. Section \ref{sect:encoder} presents this part in details.

\subsection{High-Order Attention Mechanism}
In some visual tasks, high-order relationship has been proved to be important for performance improvements.
For example, fine-grained image categorization is a task need to distinguish tiny differences between images. High-order relationship has demonstrated its effectiveness in this task. 
Lin et al. \cite{Bil2015Lin} propose bilinear models that use outer product to represent local pairwise feature interactions. Chen et al. \cite{Mix2019Chen} design a high-order attention module that utilizes the high-order polynomial predictor to capture subtle differences among objects in the image. Zhao et al. \cite{Graph2021zhao} introduce a graph-based relation discovery approach to learn positional and semantic features. When applied to RGB-D SOD, the high-order representation helps the encoder to distinguish tiny differences between two modalities. Complementary information in this visual task can be further investigated.

Besides, various forms of attention mechanisms are exploited to focus on the pivotal information in the image, and have significantly improved the performance of computer vision tasks. Hou et al. \cite{Strip} present a mixed pooling module that captures long-range dependencies and also focuses on local details simultaneously. Fu et al. \cite{DANet} design a position attention module and a channel attention module to model the semantic dependencies in spatial and channel dimensions, respectively. Yang et al. \cite{SimAM} observe the mechanism of human attention and propose a 3D attention module, which uses an energy function to calculate the attention weights. Hou et al. \cite{CANet} decompose the channel attention into two 1D feature coding processes. Inspired by these above attention mechanisms, we design the HOSF and HOCF modules to promote the interactions in cross-modality features, which will be elaborated in Section \ref{sect:HOSF} and Section \ref{sect:HOCF}. 


\section{Proposed Method}
In this section, we describe the proposed HODINet in detail.
\subsection{Overview}
\begin{figure*}[!t]
\centering
\includegraphics[width=1.00\textwidth]{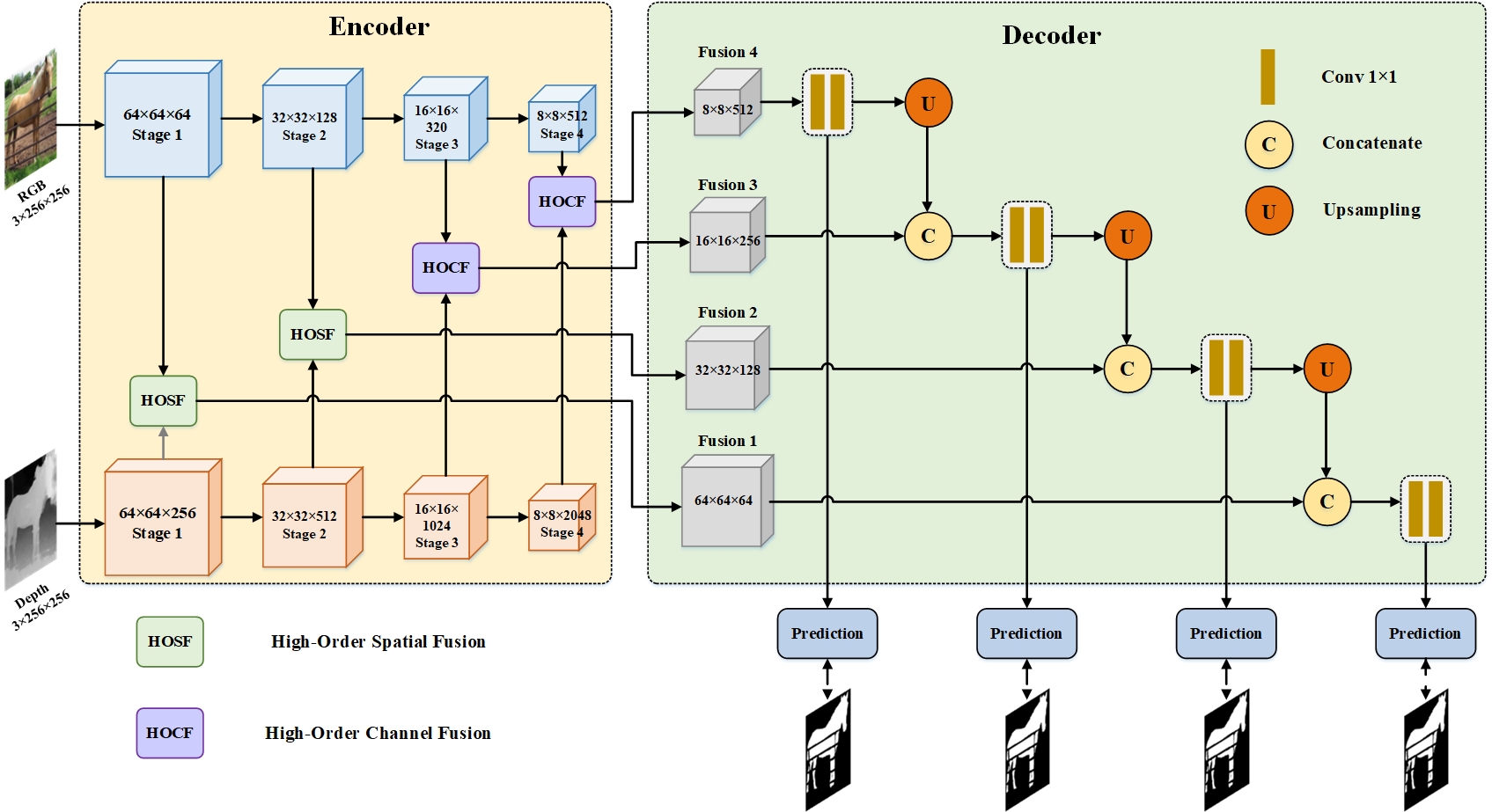}
\caption{Detailed framework of the proposed HODINet. We adopt transformer-based and CNN-based models as backbones to extract features of RGB and depth inputs, respectively. Then, the high-order spatial fusion (HOSF) module and the high-order channel fusion (HOCF) module are proposed to fuse feature maps extracted from the first two stages and the last two stages, respectively. Finally, the fused feature maps of all stages are decoded by the cascaded pyramid reconstruction network (CPRN).}
\label{fig:HODINet}
\end{figure*}

The whole framework is shown in Fig. \ref{fig:HODINet}. Given an RGB image and a corresponding depth map, we use transformer-based and CNN-based networks as backbones to extract features from RGB and depth inputs, respectively. Then, we divide the backbones into four stages
. Inspired by the recent success of high-order representation \cite{Zheng2019CVPR,Kong2017CVPR}, output features from the first two stages are fed into high-order spatial fusion (HOSF) modules to enhance the spatial characterization, while output features from the last two stages are put into high-order channel fusion (HOCF) modules to obtain the channel information. Subsequently, the fused features are input to the cascaded pyramid reconstruction network (CPRN). The CPRN progressively upsamples the resolution of feature maps to integrate the multistage information. In addition, we use deep supervision \cite{lee2015deeply} at each stage of the decoder to enhance the expression of different resolution features. The binary cross-entropy (BCE) loss is replaced by the hybrid loss \cite{8953756} for accurate boundary prediction and higher performance. Table \ref{tab:Notations} lists some operations and notations used in this paper.

\begin{table}[!t]
	\centering
	\caption{Descriptions of notations and operations used in this paper.}
	\label{tab:Notations}
	\begin{tabular}{c|c}
		\hline
		Notations & Descriptions \\ \hline
		${\rm Conv}_{k \times k}(\cdot)$ & $k$$\times$$k$ convolutional layer \\
		${\rm MO}(\cdot)$ & moment normalization \\
		${\rm Norm}(\cdot)$ & $l$-2 normalization \\
		${\rm GMP}(\cdot)$ & global max pooling \\
		${\rm GMP}^{c \times c}(\cdot)$ & global max pooling with stride of $c$$\times$$c$  \\
		${\rm GAP}(\cdot)$ & global average pooling \\
		${\rm Cat}[\cdot;\cdot]$ & concatenation \\
		${\rm FC}(\cdot)$ & full connected layer \\
		$\sigma(\cdot)$ & sigmoid activation function \\
		${\rm BN}(\cdot)$ & batch normalization (BN) \\
		${\rm ReLU}(\cdot)$ & rectified linear unit (ReLU) activation function \\
		${\rm NFE}(\cdot)$ & 1$\times$1 convolutional layer with BN and ReLU \\
		${\rm Up}_{\times s}(\cdot)$ & upsampling $s$ times with bilinear interpolations \\
		${\rm Up}_{ori}(\cdot)$ & upsampling to the original size of the input data\\\hline
		$\otimes$ & element-wise multiplication \\
		$\oplus$ & element-wise addition \\\hline
	\end{tabular}
\end{table}

\subsection{Encoder}\label{sect:encoder}

When encoding the input data of RGB stream and depth stream, we need to fully consider the differences between these two modalities. As discussed before, the RGB images contain abundant local structure information, while the depth maps intuitively describe object position relationships in the whole image. 
Therefore, we enhance the local continuity in the RGB branch by a transformer-based backbone equipped with overlapping patch embedding. 
For the depth stream, we take a stack of convolutional layers to cover the position information in a global view.


\textbf{RGB Stream.} We employ PVTv2-B3 \cite{wang2021pvtv2} as the backbone of the RGB stream, because it emphasize local continuity by overlapping patch embedding. PVTv2 can be divided into four stages when generating features of different scales. At each stage, the input image is decomposed into patches with position embedding. Then, these patches pass through a transformer-based encoder with several layers. Afterwards, the output is reshaped into feature maps for multi-stage prediction tasks. Moreover, another advantage of PVTv2 is that the resolution of output feature maps of each stage is always the same as ResNet models \cite{ResNet}, making it easy to incorporate with the CNN-based networks.

\textbf{Depth Stream.} To fully investigate the position information in a global view, we stack more convolutional layers in the backbone of the depth branch.
Since the actual receptive field is often much smaller than the theoretical ones \cite{Luo2016NeurIPS}, a deeper backbone network helps to obtain more sufficient global information. Some recent work also use deep network \cite{JL-DCF} or large convolutional kernel \cite{Zhang2021MM} to preserve large receptive field in the depth stream. In this paper, we adopt ResNet-101 \cite{ResNet} as the backbone of the depth branch.
To apply it to RGB-D SOD task, we remove the last global average pooling layer and 1000-way fully-connected layer in the original ResNet-101 network. As a result, all of the remaining blocks will be divided into four stages, expect for the first 7$\times$7 convolution and global max pooling layer. To be more specific, we set the feature maps of 11-th, 23-th, 92-th and 101-th layers as the outputs of four stages. 

\subsection{High-Order Spatial Fusion (HOSF) Module}\label{sect:HOSF}
\begin{figure*}[!t]
\centering
\includegraphics[width=1.00\textwidth]{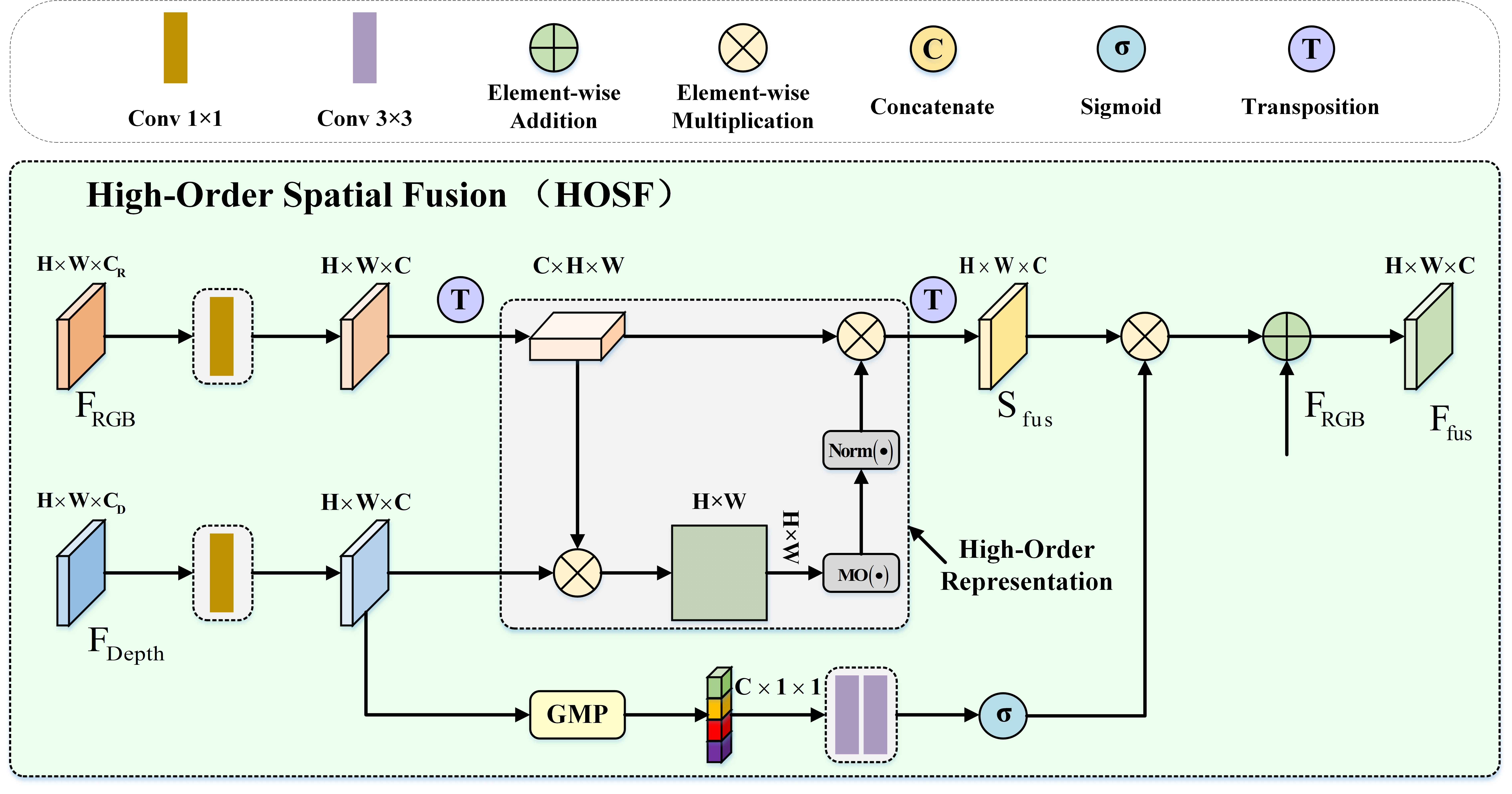}
\caption{The illustration of the proposed HOSF module.}
\label{fig:HOSF}
\end{figure*}

Complementary information between RGB images and depth maps plays a crucial role in RGB-D SOD \cite{Pang2020ECCV,Li2020ECCV}. In some visual tasks, high-order representation has shown its effectiveness to distinguish tiny differences between images. Inspired by this, we propose two high-order attention modules to further improve the performance of RGB-D SOD.

The feature maps extracted from shallower layers have a large resolution, which often contain more spatial details. Based on this, we design a High-Order Spatial Fusion (HOSF) module to integrate the RGB modality with the depth features effectively. The HOSF module utilizes the spatial high-order representation to enhance complementary information. The complete architecture of HOSF is illustrated in Fig. \ref{fig:HOSF}.

Specifically, we obtain the RGB features $\mathcal{F}^{RGB}_{i}$ and the depth features $\mathcal{F}^{Depth}_{i}$ from their corresponding backbones, where $i\in \left \{ 1,2 \right \} $ indicates the stage indexes of the encoder. Since outputs of different backbones are inconsistent in channel dimension, we first apply a 1$\times$1 convolutional layer to align the number of channels,
\begin{align}
f^{RGB}_i = &{\rm ReLU}({\rm BN}({\rm Conv}_{1\times1}(\mathcal{F}^{RGB}_i))),\\
f^{Depth}_i = &{\rm ReLU}({\rm BN}({\rm Conv}_{1\times1}(\mathcal{F}^{Depth}_i))),
\end{align}
where ${\rm Conv}_{1\times1}(\cdot)$ denotes a 1$\times$1 convolutional layer, ${\rm BN}(\cdot)$ represents a batch normalization (BN) layer \cite{ioffe2015batch} and ${\rm ReLU}(\cdot)$ symbolizes a rectified linear unit (ReLU) activation function.

Then, we extract high-order representations \cite{zhao2021graph} to jointly learn the spatial relationship between $f^{RGB}_i$ and $f^{Depth}_i$,
\begin{equation}{
\mathcal{A}^{sp}_i={\rm Norm}({\rm MO}((f^{RGB}_i)^T \otimes f^{Depth}_i)) \otimes (f^{RGB}_i)^T,
}
\end{equation}
where $T$ stands for the transposition operation, $\otimes$ indicates the element-wise multiplication operation, ${\rm MO}(x)=sign(x) \cdot x^{-1/2}$ denotes the moment, and ${\rm Norm}(x)=x/\left \| x \right \|  _{2}^{2} $ represents the $l$-2 normalization.

Depth data is generally obtained by depth sensors. The foreground and background have different depth values because of their distances to the sensor, which allows backbone networks to learn complementary information apart from RGB data. However, not all depth information can bring beneficial improvements to the RGB-D SOD task. The low-quality depth maps will introduce additional noise to the predicted saliency maps \cite{SIP_D3Net2021TNNLS,9686679}. To fuse the depth maps effectively, we integrate the depth information selectively by using a global max pooling operation followed by two 3$\times$3 convolutional layers, which can be expressed as,
\begin{equation}
f^{DW}_i = {\rm \sigma}({\rm Conv}_{3 \times 3}({\rm Conv}_{3 \times 3}({\rm GMP}(f^{Depth}_i)))),
\end{equation}
where ${\rm GMP}=(\cdot)$ denotes the global max pooling operation, $\sigma(\cdot)$ represents the sigmoid activation function. Then, we supplement the depth clues to the fused features by element-wise multiplication operation. Finally, a residual connection is used to retain the global information in RGB stream. The final fused feature maps is formulated as:
\begin{equation}
\begin{split}
\mathcal{F}^{fus}_i = (\mathcal{A}^{sp}_i \otimes f^{DW}_i) \oplus f^{RGB}_i,\\
i \in \{1,2\},
\end{split}
\end{equation}
where $\oplus$ represents the element-wise addition operation.

The HOSF module is proposed to interact the feature maps in shallow layers of the encoders. In the next subsection, we will introduce anther high-order attention module for deep layers in the backbones.

\subsection{High-Order Channel Fusion (HOCF) Module}\label{sect:HOCF}
\begin{figure*}[!t]
	\centering
	\includegraphics[width=1.00\textwidth]{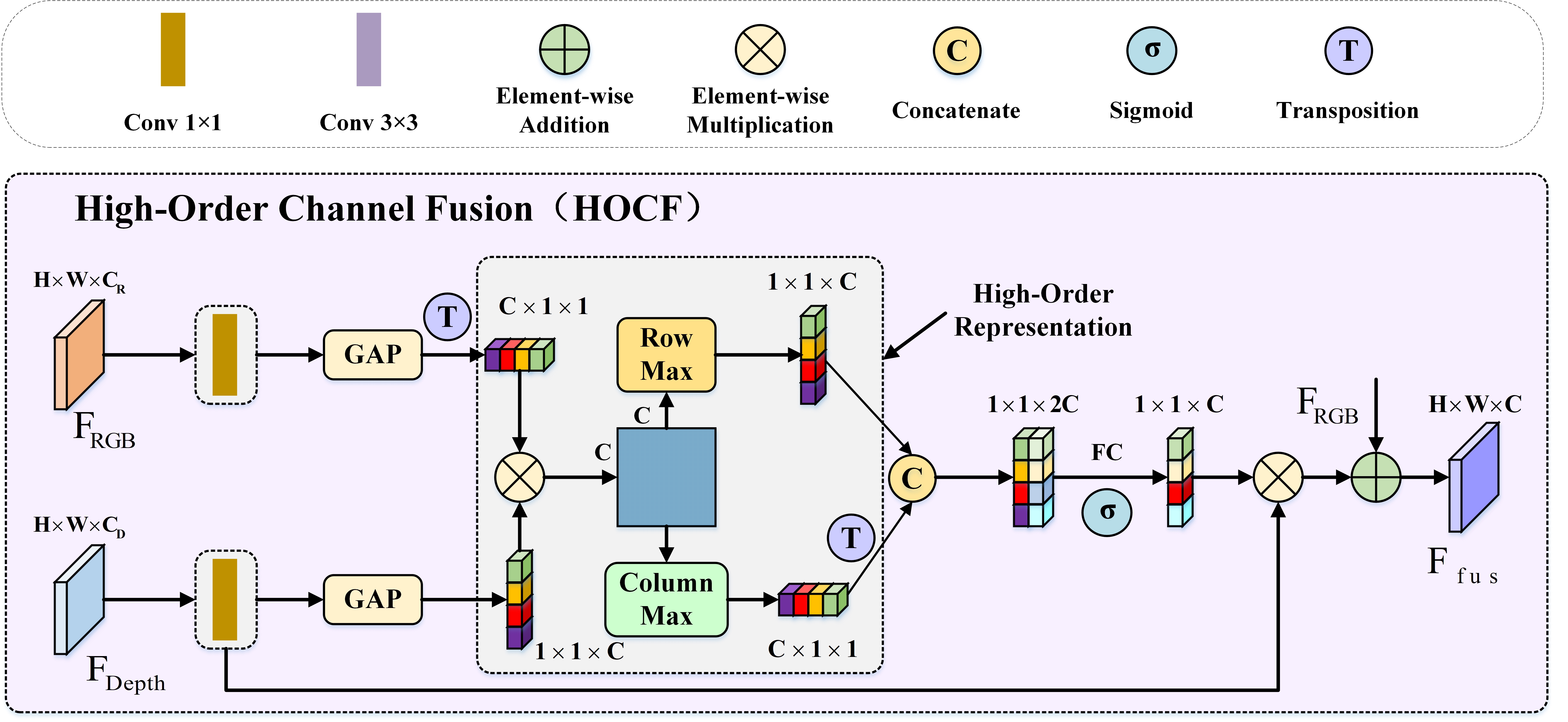}
	\caption{The illustration of the proposed HOCF module.}
	\label{fig:HOCF}
\end{figure*}
In encoder networks, the features extracted from deep layers have more channels than the features in shallow layers. Each channel represents a type of semantic information. By modeling the channel interactions in a high-order manner, we can emphasize complementary feature representation for specific semantics. Thus, we introduce the High-Order Channel Fusion (HOCF) module to exploit the channel relationships between two modalities. The illustration of HOCF is shown in Fig. \ref{fig:HOCF}.

In the HOCF module, we first adopt a 1$\times$1 convolutional layer to align the number of channels,
\begin{align}
f^{RGB}_i = &{\rm ReLU}({\rm BN}({\rm Conv}_{1\times1}(\mathcal{F}^{RGB}_i))),\\
f^{Depth}_i = &{\rm ReLU}({\rm BN}({\rm Conv}_{1\times1}(\mathcal{F}^{Depth}_i))),
\end{align}
where $i\in \left \{ 3,4 \right \} $ is the indexes of stage in encoder backbones. Then, the global average pooling operation is added to generate the corresponding channel attention mask. We employ the element-wise multiplication operation to calculate the interactive channel attention map $\mathcal{A}^{ch}_i \in \mathbb{R}^{C\times C}$,
\begin{align}
\mathcal{A}^{RGB}_i =& {\rm GAP}(f^{RGB}_i),\\
\mathcal{A}^{Depth}_i =& {\rm GAP}(f^{Depth}_i),\\
\mathcal{A}^{ch}_i =& \mathcal{A}^{RGB}_i \otimes \mathcal{A}^{Depth}_i.
\end{align}
After these operations, we obtain the maximum value of each row and each column in the interactive channel attention map, respectively. Then, we apply a series of operations to further integrate the most prominent features, 
\begin{align}
\tilde{\mathcal{A}}^{ch}_i = & \sigma({\rm FC}({\rm Cat}[{\rm GMP}^{C \times 1}(\mathcal{A}^{ch}_i);{\rm GMP}^{1 \times C}(\mathcal{A}^{ch}_i)]),
\end{align}
where ${\rm Cat}[\cdot,\cdot]$ denotes a concatenation operation, ${\rm FC}(\cdot)$ stands for a fully connected (FC) layer, $\sigma(\cdot)$ is a sigmoid activation function. Here, ${\rm GMP}^{C \times 1}(\mathcal{A}^{ch}_i)$ calculates how each channel of RGB feature interacts with the depth modality, and ${\rm GMP}^{1 \times C}(\mathcal{A}^{ch}_i)$ obtains how each channel of depth feature correlates with the RGB modality. Finally, we use element-wise multiplication operation and residual connection to obtain fused feature maps,
\begin{equation}
\begin{split}
\mathcal{F}^{fus}_i = (\tilde{\mathcal{A}}^{ch}_i\otimes f^{Depth}_i) \oplus f^{RGB}_i,\\
i \in\{3,4\}.
\end{split}
\end{equation}
These feature maps are then fed into the decoder to recover the resolutions.

\subsection{Decoder}
In the decoder, if we upsample the feature maps directly to the raw size of the inputs, it will produce excessive noise and imprecise results \cite{liu2021tritransnet}. This is unacceptable for pixel-level prediction tasks. As mentioned in \cite{7780576}, combining upsample operations with convolutional layers is crucial for restoring the resolution in decoding process. Inspired by the principle of PVT \cite{wang2021pyramid} and FPN \cite{lin2017feature}, we employ a cascaded pyramid reconstruction network (CPRN) to gradually integrate multi-level features via a top-down pathway. Given the output $\mathcal{F}^{fus}_i$, $i \in\{1,2,3,4\}$, from four high-order fusion modules, we first apply two 1$\times$1 convolutional layers to reduce the number of channels. Then, the feature maps are upsampled two times with bilinear interpolations to make the resolution compatible to next stage. After that, we concatenate these feature maps. The above operations are continued until the feature map is restored to the size of original inputs. 

For brevity, we first denote a nonlinear feature enhancement (NFE) unit before introducing the cascaded pyramid reconstruction network,
\begin{equation}
{\rm NFE}(\cdot)={\rm ReLU}({\rm BN}({\rm Conv}_{1\times1}(\cdot))).
\end{equation} 
Thus, we use the following formulas to present the process of cascaded pyramid decoder,
\begin{align}
\mathcal{F}_4^{out} =&{\rm NFE}({\rm NFE}(\mathcal{F}^{fus}_4)),\\
\mathcal{F}_3^{out} =&{\rm NFE}({\rm NFE}({\rm Cat}(\mathcal{F}^{fus}_3;{\rm Up}_{\times 2}(\mathcal{F}_4^{out})))),\\
\mathcal{F}_2^{out} =&{\rm NFE}({\rm NFE}({\rm Cat}(\mathcal{F}^{fus}_2;{\rm Up}_{\times 2}(\mathcal{F}_3^{out})))),\\
\mathcal{F}_1^{out} =&{\rm NFE}({\rm NFE}({\rm Cat}(\mathcal{F}^{fus}_1;{\rm Up}_{\times 2}(\mathcal{F}_2^{out})))).
\end{align}
Meanwhile, we add deep supervisions \cite{DS} to the outputs $\mathcal{F}^{out}_i$, $i \in\{1,2,3,4\}$, of the cascaded pyramid decoder to speed up convergence. The predicted saliency maps can be formulated as,
\begin{equation}
\begin{split}
P_i={\rm Up}_{ori}({\rm Conv}_{1 \times 1}({\rm ReLU}({\rm BN}({\rm Conv}_{3 \times 3}(\mathcal{F}^{out}_i))))),\\
i \in\{1,2,3,4\},
\end{split}
\end{equation}
where $P_i$ represents the predictions in the $i$-th stage, ${\rm Up}_{ori}(\cdot)$ denotes upsampling the feature maps to the original resolution of the input image. Note that only $P_1$ is used as the final saliency map, while other three predictions are omitted in the test phase.

\subsection{Loss Function}
For each of the decoder stage, we adopt the hybrid loss function \cite{8953756} as the loss function, which is a combination of binary cross-entropy (BCE) loss \cite{BCE}, SSIM loss \cite{SSIM} and IoU loss \cite{IoU}. It can be defined as below, 
\begin{equation}
\begin{split}
\mathcal{L}_i^{hybrid} = \mathcal{L}_i^{BCE} + \mathcal{L}_i^{SSIM} + \mathcal{L}_i^{IoU},\\
i \in \left\{ {1,2,3,4} \right\}.
\end{split}
\end{equation}
BCE loss measures the pixel-level difference between the predicted saliency map $P\in\left [0,1 \right]^{W \times H}$ and ground truth $G\in\left \{ 0,1 \right \} ^{W\times H}$, which can be denoted as,
\begin{equation}
\begin{split}
& \mathcal{L}^{BCE} = \\
& - \sum\limits_{w,h=1}^{W,H} {[{G_{wh}}\log({P_{wh}}) + (1 - {G_{wh}})\log(1 - {P_{wh}})]}.
\end{split}
\end{equation}

SSIM loss evaluates the structural similarity, which is defined as, 
\begin{equation}
\mathcal{L}^{SSIM} = 1 - \frac{{(2{\mu _P}{\mu _G} + {C_1})(2{\sigma _{PG}} + {C_2})}}{{(\mu _P^2 + \mu _G^2 + {C_1})(\sigma _P^2 + \sigma _G^2 + {C_2})}},
\end{equation}
where $\mu _P, \mu _G$ and $\sigma _P, \sigma _G$ are the means and the standard deviations of the two patches from the predicted saliency map $P$ and the ground truth $G$, respectively. $\sigma _{PG}$ is their covariance. Additionally, $C_1 = 0.01^2$ and $C_2 = 0.03^2$ are set to avoid the numerator being divided by zero. 

IoU loss is adopted to capture fine structures in image level, which is computed as,
\begin{equation}
\mathcal{L}^{IoU} = 1 -  {\frac{\sum\limits_{w,h=1}^{W,H} {P_{wh}}{G_{wh}}} {\sum\limits_{w,h=1}^{W,H} ({P_{wh}} + {G_{wh}} - {P_{wh}}{G_{wh}} )}}.
\end{equation}
As a result, the total loss can be represented as,
\begin{equation}
\begin{split}
\mathcal{L}_{total}= \sum\limits_{i=1}^{4} {{\mathcal{L}_i^{hybrid}}({P_i},G)},\\
i \in \{1,2,3,4\}.
\end{split}
\end{equation}
All the predicted saliency maps and the ground truths have the same resolution as original input images.

\section{Discussion}
\subsection{The novelty of the proposed HOSF and HOCF}
The key of RGB-D SOD lies in the complementary information between the two modalities, and high-order information has been proven to distinguish subtle differences between images in some other computer vision tasks. Although high-order attention mechanism has been proposed for many years, it is often only adopted for single modality feature enhancement. The cross-modality higher-order attention obtains relatively less research interests. To the best of our knowledge, we are the first to use higher-order statistical information for this cross-modality vision task, namely RGB-D SOD. As we can see in the ablation experiment in Section \ref{Sect:Abl}, HOSF and HOCF actually improve the performance.

\subsection{The choices of backbones in RGB branch and depth branch}
In this paper, we employ PVTv2 and ResNet-101 as the backbones for the RGB branch and the depth branch, respectively. In the preliminary experiment, we found that this implementation achieved the best results. The reason may be that RGB images contain more valuable information. If both streams use PVTv2 as encoder identically, there will be a slight decrease in performance. Compared with other transformer-based SOD methods \cite{TriTransNet}, we still achieved competitive results. In other fields of computer vision, more advanced backbone networks are naturally used to improve the performance \cite{Guo2022CVPR}. 

\section{Experiments}
In this section, we first introduce the experimental setup and implementation details. Then, we compare the proposed HODINet with other state-of-the-art RGB-D SOD methods on seven benchmark datasets and analyse the experimental results. The ablation studies are also conducted to verify the effectiveness of each component. Finally, some failure cases are discussed.

\begin{table*}[!t]
    \centering
    \caption{Quantitative results compared with state-of-the-art RGB-D SOD methods. ``-'' means that the results are unavailable since the authors did not release them. $\uparrow$ ($\downarrow$) indicates the larger (smaller), the better. The best and the second best results are highlighted in bold and underline, respectively.}
    \label{tab:quantitative}
\resizebox{\linewidth}{!}{
\begin{tabular}{c|cc|cc|cc|cc|cc|cc|cc}
\hline
 & \multicolumn{2}{c|}{NLPR \cite{NLPR}} & \multicolumn{2}{c|}{DUT-RGBD \cite{DUT-RGBD}} & \multicolumn{2}{c|}{NJUD \cite{NJUD}} & \multicolumn{2}{c|}{STEREO \cite{STEREO}} & \multicolumn{2}{c|}{SSD \cite{SSD}} & \multicolumn{2}{c|}{LFSD \cite{LFSD}} & \multicolumn{2}{c}{SIP \cite{SIP_D3Net2021TNNLS}} \\
\multirow{-2}{*}{Model} & $S_\alpha$ $\uparrow$ & MAE $\downarrow$ & $S_\alpha$ $\uparrow$ & MAE $\downarrow$ & $S_\alpha$ $\uparrow$ & MAE $\downarrow$ & $S_\alpha$ $\uparrow$ & MAE $\downarrow$ & $S_\alpha$ $\uparrow$ & MAE $\downarrow$ & $S_\alpha$ $\uparrow$ & MAE $\downarrow$ & $S_\alpha$ $\uparrow$ & MAE $\downarrow$ \\ \hline
Ours & \textbf{0.934} & \textbf{0.018} & \textbf{0.942} & \textbf{0.022} & \textbf{0.925} & \textbf{0.030} & \textbf{0.917} & \textbf{0.033} & \textbf{0.885} & \textbf{0.038} & \textbf{0.886} & \textbf{0.054} & \textbf{0.900} & \textbf{0.039} \\ \hline
SMAC(21TPAMI)\cite{SMAC} & 0.922 & 0.027 & 0.926 & 0.033 & 0.903 & 0.044 & 0.905 & 0.042 & {\ul 0.884} & {\ul 0.044} & 0.875 & {\ul 0.063} & 0.883 & {\ul 0.049} \\
CDINet(21MM)\cite{Zhang2021MM} & 0.927 & 0.024 & 0.926 & 0.030 & 0.918 & 0.036 & 0.905 & 0.041 & 0.852 & 0.057 & 0.870 & 0.064 & 0.875 & 0.055 \\
DCF(21CVPR)\cite{DCF} & 0.921 & 0.023 & 0.924 & 0.031 & 0.903 & 0.039 & 0.905 & {\ul 0.037} & 0.851 & 0.054 & 0.855 & 0.071 & 0.873 & 0.052 \\
DSA2F(21CVPR)\cite{DSA2F} & 0.918 & 0.024 & 0.921 & 0.030 & 0.903 & 0.039 & 0.904 & 0.038 & 0.876 & 0.045 & {\ul 0.882} & \textbf{0.054} & 0.860 & 0.057 \\
DQSD(21TIP)\cite{DQSD} & 0.915 & 0.030 & 0.844 & 0.073 & 0.898 & 0.051 & 0.891 & 0.052 & 0.868 & 0.053 & 0.850 & 0.085 & 0.863 & 0.065 \\
DRLF(21TIP)\cite{DRLF} & 0.902 & 0.032 & 0.825 & 0.080 & 0.886 & 0.055 & 0.888 & 0.050 & 0.834 & 0.066 & 0.834 & 0.090 & 0.850 & 0.071 \\
HAINet(21TIP)\cite{HAINet} & 0.821 & 0.025 & 0.910 & 0.038 & 0.909 & 0.038 & 0.909 & 0.038 & 0.857 & 0.053 & 0.853 & 0.080 & {\ul 0.886} & 0.048 \\
CDNet(21TIP)\cite{CDNet} & 0.927 & 0.025 & 0.927 & 0.031 & 0.918 & 0.036 & 0.906 & 0.040 & 0.876 & 0.045 & 0.858 & 0.073 & 0.874 & 0.054 \\
RD3D(21AAAI)\cite{RD3D} & {\ul 0.930} & 0.022 & {\ul 0.932} & 0.031 & 0.916 & 0.036 & 0.911 & {\ul 0.037} & - & - & 0.857 & 0.074 & 0.885 & 0.048 \\
DRSD(21TMM)\cite{DRSD} & 0.906 & 0.038 & 0.864 & 0.072 & 0.886 & 0.050 & 0.899 & 0.046 & 0.861 & 0.049 & 0.827 & 0.092 & - & - \\
S3Net(21TMM)\cite{S3Net} & 0.927 & {\ul 0.021} & 0.912 & 0.035 & 0.913 & {\ul 0.034} & {\ul 0.913} & 0.038 & - & - & 0.874 & 0.066 & 0.875 & 0.051 \\
ATSA(20ECCV)\cite{ATSA} & 0.909 & 0.027 & 0.916 & 0.033 & 0.885 & 0.047 & 0.896 & 0.038 & 0.820 & 0.052 & 0.845 & 0.078 & 0.849 & 0.063 \\
CoNet(20ECCV)\cite{CoNet} & 0.912 & 0.027 & 0.923 & {\ul 0.029} & 0.896 & 0.046 & 0.905 & {\ul 0.037} & 0.851 & 0.056 & 0.848 & 0.076 & 0.860 & 0.058 \\
DANet(20ECCV)\cite{DANet2020ECCV} & 0.920 & 0.027 & 0.899 & 0.042 & 0.899 & 0.046 & 0.901 & 0.044 & 0.864 & 0.050 & 0.841 & 0.087 & 0.875 & 0.055 \\
BBSNet(20ECCV)\cite{BBSNet_Fan2020ECCV} & 0.931 & 0.023 & 0.882 & 0.058 & {\ul 0.921} & 0.035 & 0.908 & 0.041 & 0.863 & 0.052 & 0.835 & 0.092 & 0.879 & 0.055 \\
A2dele(20CVPR)\cite{A2dele2020CVPR} & 0.899 & 0.029 & 0.885 & 0.043 & 0.871 & 0.051 & 0.879 & 0.045 & 0.803 & 0.070 & 0.825 & 0.084 & 0.829 & 0.070 \\
SSF(20CVPR)\cite{SSF2020CVPR} & 0.915 & 0.027 & 0.915 & 0.033 & 0.899 & 0.043 & 0.837 & 0.065 & 0.790 & 0.084 & 0.851 & 0.074 & 0.799 & 0.091 \\
UCNet(20CVPR)\cite{Zhang2020CVPR} & 0.920 & 0.025 & 0.871 & 0.059 & 0.897 & 0.043 & 0.903 & 0.039 & 0.865 & 0.049 & 0.856 & 0.074 & 0.875 & 0.051 \\
S2MA(20CVPR)\cite{Liu2020CVPR} & 0.915 & 0.030 & 0.903 & 0.043 & 0.894 & 0.053 & 0.890 & 0.051 & 0.868 & 0.052 & 0.837 & 0.094 & 0.872 & 0.058 \\
JLDCF(20CVPR)\cite{JL-DCF} & 0.925 & 0.022 & 0.906 & 0.042 & 0.902 & 0.041 & 0.903 & 0.040 & 0.860 & 0.053 & 0.853 & 0.077 & 0.880 & 0.049 \\
D3Net(20TNNLS)\cite{SIP_D3Net2021TNNLS} & 0.912 & 0.030 & 0.850 & 0.071 & 0.900 & 0.046 & 0.899 & 0.046 & 0.857 & 0.058 & 0.825 & 0.095 & 0.860 & 0.063 \\
DMRA(19ICCV)\cite{DUT_DMRA2019ICCV} & 0.899 & 0.031 & 0.889 & 0.048 & 0.886 & 0.051 & 0.886 & 0.047 & 0.857 & 0.058 & 0.847 & 0.075 & 0.806 & 0.085 \\
CPFP(19CVPR)\cite{CPFP} & 0.888 & 0.036 & 0.818 & 0.076 & 0.878 & 0.053 & 0.879 & 0.051 & 0.807 & 0.082 & 0.828 & 0.088 & 0.850 & 0.064 \\
TANet(19TIP)\cite{TANet} & 0.886 & 0.041 & 0.808 & 0.093 & 0.878 & 0.060 & 0.871 & 0.060 & 0.839 & 0.063 & 0.801 & 0.111 & 0.835 & 0.075 \\ \hline
\hline
 & \multicolumn{2}{c|}{NLPR \cite{NLPR}} & \multicolumn{2}{c|}{DUT-RGBD \cite{DUT-RGBD}} & \multicolumn{2}{c|}{NJUD \cite{NJUD}} & \multicolumn{2}{c|}{STEREO \cite{STEREO}} & \multicolumn{2}{c|}{SSD \cite{SSD}} & \multicolumn{2}{c|}{LFSD \cite{LFSD}} & \multicolumn{2}{c}{SIP \cite{SIP_D3Net2021TNNLS}} \\
\multirow{-2}{*}{Model} & $E_\xi$ $\uparrow$ & $F_\beta$ $\uparrow$ & $E_\xi$ $\uparrow$ & $F_\beta$ $\uparrow$ & $E_\xi$ $\uparrow$ & $F_\beta$ $\uparrow$ & $E_\xi$ $\uparrow$ & $F_\beta$ $\uparrow$ & $E_\xi$ $\uparrow$ & $F_\beta$ $\uparrow$ & $E_\xi$ $\uparrow$ & $F_\beta$ $\uparrow$ & $E_\xi$ $\uparrow$ & $F_\beta$ $\uparrow$ \\ \hline
Ours & \textbf{0.966} & \textbf{0.933} & \textbf{0.973} & \textbf{0.958} & \textbf{0.952} & \textbf{0.935} & \textbf{0.953} & \textbf{0.924} & {\ul 0.926} & \textbf{0.891} & \textbf{0.925} & {\ul 0.899} & \textbf{0.942} & \textbf{0.928} \\ \hline
SMAC(21TPAMI)\cite{SMAC} & 0.953 & 0.904 & 0.956 & 0.928 & 0.937 & 0.896 & 0.941 & 0.897 & \textbf{0.928} & 0.869 & 0.911 & 0.870 & 0.925 & 0.886 \\
CDINet(21MM)\cite{Zhang2021MM} & 0.953 & 0.923 & 0.956 & 0.944 & 0.944 & 0.827 & - & 0.903 & 0.906 & 0.867 & {\ul 0.914} & 0.889 & 0.911 & {\ul 0.903} \\
DCF(21CVPR)\cite{DCF} & 0.956 & 0.917 & 0.956 & 0.940 & 0.940 & 0.917 & 0.943 & 0.914 & 0.905 & 0.857 & 0.902 & 0.878 & 0.920 & 0.899 \\
DSA2F(21CVPR)\cite{DSA2F} & 0.950 & 0.916 & 0.950 & 0.939 & 0.923 & 0.918 & 0.933 & 0.911 & 0.904 & {\ul 0.877} & 0.923 & \textbf{0.903} & 0.911 & 0.891 \\
DQSD(21TIP)\cite{DQSD} & 0.934 & 0.909 & 0.889 & 0.859 & 0.912 & 0.910 & 0.911 & 0.900 & 0.890 & {\ul 0.877} & 0.883 & 0.875 & 0.900 & 0.890 \\
DRLF(21TIP)\cite{DRLF} & 0.935 & 0.904 & 0.870 & 0.851 & 0.901 & 0.883 & 0.915 & 0.878 & 0.879 & 0.859 & 0.872 & 0.857 & 0.891 & 0.868 \\
HAINet(21TIP)\cite{HAINet} & 0.954 & 0.908 & 0.944 & 0.920 & 0.941 & 0.909 & {\ul 0.947} & 0.909 & 0.906 & 0.859 & 0.892 & 0.876 & 0.927 & {\ul 0.903} \\
CDNet(21TIP)\cite{CDNet} & 0.955 & {\ul 0.926} & 0.956 & {\ul 0.945} & {\ul 0.950} & 0.919 & 0.942 & 0.898 & 0.924 & 0.871 & 0.896 & 0.861 & 0.917 & 0.902 \\
RD3D(21AAAI)\cite{RD3D} & {\ul 0.965} & 0.919 & {\ul 0.960} & 0.939 & 0.947 & 0.914 & {\ul 0.947} & 0.906 & - & - & 0.897 & 0.878 & 0.924 & 0.889 \\
DRSD(21TMM)\cite{DRSD} & 0.936 & 0.882 & 0.902 & 0.853 & 0.927 & 0.876 & 0.933 & 0.887 & 0.917 & 0.832 & 0.866 & 0.813 & - & - \\
S3Net(21TMM)\cite{S3Net} & 0.962 & 0.923 & 0.939 & 0.922 & 0.944 & {\ul 0.928} & 0.945 & {\ul 0.918} & - & - & 0.902 & 0.892 & {\ul 0.933} & 0.891 \\
ATSA(20ECCV)\cite{ATSA} & 0.951 & 0.898 & 0.953 & 0.928 & 0.930 & 0.893 & 0.942 & 0.901 & 0.920 & 0.853 & 0.893 & 0.859 & 0.901 & 0.861 \\
CoNet(20ECCV)\cite{CoNet} & 0.948 & 0.893 & 0.959 & 0.932 & 0.937 & 0.893 & {\ul 0.947} & 0.901 & 0.917 & 0.837 & 0.895 & 0.852 & 0.917 & 0.873 \\
DANet(20ECCV)\cite{DANet2020ECCV} & 0.955 & 0.909 & 0.939 & 0.904 & 0.935 & 0.898 & 0.937 & 0.892 & 0.914 & 0.843 & 0.874 & 0.840 & 0.918 & 0.876 \\
BBSNet(20ECCV)\cite{BBSNet_Fan2020ECCV} & 0.961 & 0.918 & 0.912 & 0.870 & 0.949 & 0.919 & 0.942 & 0.903 & 0.914 & 0.843 & 0.870 & 0.828 & 0.922 & 0.884 \\
A2dele(20CVPR)\cite{A2dele2020CVPR} & 0.944 & 0.882 & 0.928 & 0.891 & 0.916 & 0.874 & 0.928 & 0.880 & 0.862 & 0.777 & 0.866 & 0.828 & 0.890 & 0.834 \\
SSF(20CVPR)\cite{SSF2020CVPR} & 0.953 & 0.896 & 0.950 & 0.923 & 0.935 & 0.896 & 0.912 & 0.840 & 0.867 & 0.765 & 0.892 & 0.863 & 0.870 & 0.786 \\
UCNet(20CVPR)\cite{Zhang2020CVPR} & 0.956 & 0.903 & 0.908 & 0.864 & 0.936 & 0.895 & 0.944 & 0.899 & 0.907 & 0.855 & 0.898 & 0.860 & 0.919 & 0.879 \\
S2MA(20CVPR)\cite{Liu2020CVPR} & 0.953 & 0.902 & 0.937 & 0.901 & 0.930 & 0.889 & 0.932 & 0.882 & 0.909 & 0.848 & 0.873 & 0.835 & 0.919 & 0.877 \\
JLDCF(20CVPR)\cite{JL-DCF} & 0.963 & 0.918 & 0.941 & 0.910 & 0.944 & 0.904 & {\ul 0.947} & 0.904 & 0.902 & 0.833 & 0.894 & 0.863 & 0.925 & 0.889 \\
D3Net(20TNNLS)\cite{SIP_D3Net2021TNNLS} & 0.953 & 0.897 & 0.889 & 0.842 & 0.939 & 0.900 & 0.938 & 0.891 & 0.910 & 0.834 & 0.862 & 0.810 & 0.909 & 0.861 \\
DMRA(19ICCV)\cite{DUT_DMRA2019ICCV} & 0.947 & 0.879 & 0.933 & 0.898 & 0.927 & 0.886 & 0.938 & 0.886 & 0.906 & 0.844 & 0.900 & 0.856 & 0.875 & 0.821 \\
CPFP(19CVPR)\cite{CPFP} & 0.932 & 0.863 & 0.859 & 0.795 & 0.923 & 0.877 & 0.925 & 0.874 & 0.852 & 0.766 & 0.872 & 0.826 & 0.903 & 0.851 \\
TANet(19TIP)\cite{TANet} & 0.941 & 0.863 & 0.861 & 0.790 & 0.925 & 0.874 & 0.923 & 0.861 & 0.897 & 0.810 & 0.847 & 0.796 & 0.895 & 0.830 \\ \hline
\end{tabular}
}
\end{table*}

\begin{figure*}[!t]
	\centering
	\includegraphics[width=1.00\textwidth]{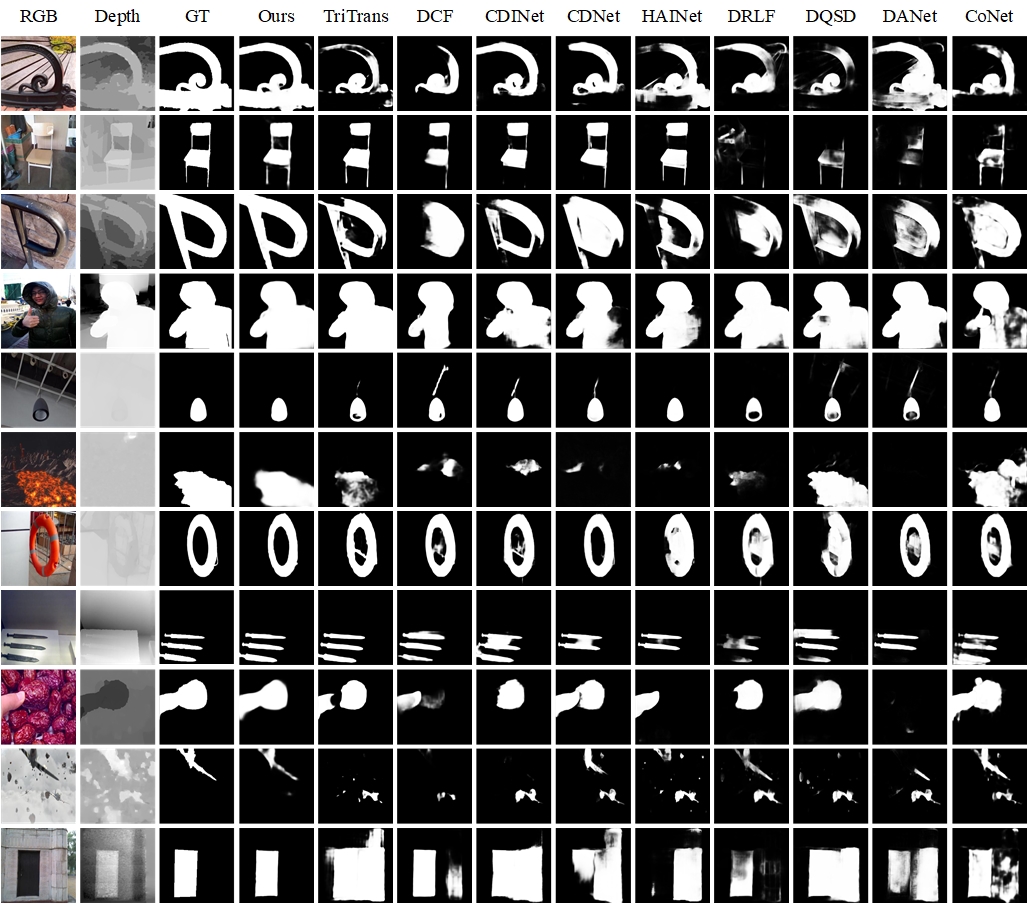}
	\centering
	\caption{Visual comparisons of the proposed HODINet and other state-of-the-art RGB-D SOD methods. Our approach obtain competitive performance in a variety of complex scenarios.}
	\label{fig:qualitative}
\end{figure*}

\subsection{Datasets and Evaluation Metrics}
\textbf{Datasets.} To demonstrate the generalization capability of the proposed model, we conduct experiments on seven publicly available datasets. 
\textbf{NLPR} \cite{NLPR} includes 1,000 stereo images from 11 types of indoor and outdoor scenes.
\textbf{DUT-RGBD} \cite{DUT-RGBD} contains 1,200 images captured by Lytro Illum camera. 
\textbf{NJUD} \cite{NJUD} is composed of 1,985 image pairs from 3D movies and photographs. 
\textbf{STEREO} \cite{STEREO} collects 1,000 pairs of binocular images from the Internet. 
\textbf{SSD} \cite{SSD} and \textbf{LFSD} \cite{LFSD} are two small-scale datasets with only 80 and 100 image pairs, respectively. \textbf{SIP} \cite{SIP_D3Net2021TNNLS} involves 929 images with high quality depth maps. It covers many challenging scenes of salient objects and persons. 
Following the same settings as previous work \cite{SSF2020CVPR,DCF,Zhang2021MM}, we use a combined set of three datasets to train the proposed model, including 700 images from NLPR, 800 pairs from DUT-RGBD and 1,485 samples from NJUD. The remaining images with corresponding depth maps in these datasets are used for test. 

\textbf{Evaluation metrics.} We adopt four widely used metrics to comprehensively compare our HODINet with other methods, including mean absolute error (MAE) \cite{Borji2015TIP}, S-measure ($S_\alpha$) \cite{S-measure}, max F-measure ($F_\beta$) \cite{Achanta2009CVPR} and max E-measure ($E_\xi$) \cite{Fan2018IJCAI}. MAE calculates the mean absolute difference in pixel-level between the predicted saliency map and the ground truth. The S-measure is designed to evaluate the structural similarities in object-level and region-level. Both F-measure and E-measure first obtain the binary saliency maps by varying the thresholds. The max F-measure computes the harmonic mean of average precision and average recall in multiple thresholds, while E-measure utilizes global image-level and local pixel-level information to measure the converted binary maps.

\subsection{Implementation Details}
We conduct all experiments with an NVIDIA GeForce RTX 2080Ti GPU on the Pytorch framework. The transformer-based and CNN-based backbones are pretrained on ImageNet \cite{deng2009imagenet}. The other layers are initialized to the default settings in Pytorch. During the training and test phase, we simply copy the input depth maps to three channels and resize all images to 256$\times$256. Besides, multiple data augmentations, such as random cropping, flipping, rotating and color enhancement, are used in the training phase to avoid overfitting. We use the Adam optimizer \cite{ADAM} to train the proposed network in an end-to-end manner with a batch size of 4 for 100 epochs. The initial learning rate is 1e-4 and then decreases by a factor of 0.9 every epoch. The whole training procedure takes about 5 hours for convergence. The inference speed is around 35 FPS when testing an image with a size of 256$\times$256 on a single NVIDIA RTX 2080Ti. \label{experimental_setup}

\subsection{Comparison With State-of-the-art Methods}
In this subsection, we compare the proposed HODINet with 24 state-of-the-art RGB-D SOD methods, including SMAC \cite{SMAC}, CDINet \cite{Zhang2021MM}, DCF \cite{DCF}, DSA2F \cite{DSA2F}, DQSD \cite{DQSD}, DRLF \cite{DRLF}, HAINet \cite{HAINet}, CDNet \cite{CDNet}, RD3D \cite{RD3D}, DRSD \cite{DRSD}, S3Net \cite{S3Net}, ATSA \cite{ATSA}, CoNet \cite{CoNet}, DANet \cite{DANet2020ECCV}, BBSNet \cite{BBSNet_Fan2020ECCV}, A2dele \cite{A2dele2020CVPR}, SSF \cite{SSF2020CVPR}, UCNet \cite{Zhang2020CVPR}, S2MA \cite{Liu2020CVPR}, JLDCF \cite{JL-DCF}, D3Net \cite{SIP_D3Net2021TNNLS}, DMRA \cite{DUT_DMRA2019ICCV}, CPFP \cite{CPFP}, and TANet \cite{TANet}. We directly report the quantitative results in the published papers if accessible. For the other baselines, we calculate the evaluation metrics by the saliency maps provided by the authors. All the metrics are calculated by the official evaluation tools.

\subsubsection{Quantitative Analysis}
Table. \ref{tab:quantitative} illustrates the quantitative results of our method. We calculate four evaluation metrics on seven datasets for comprehensive comparisons. Higher values of $S_{\alpha}$, $F_{\beta}$ and $E_{\xi}$ indicate better performance. On the contrary, ${\rm MAE}$ is the opposite. We place other baselines in reverse chronological order from top to bottom. Compared with other state-of-the-art methods, the proposed HODINet obtains comparable performance. 
Concretely, the results of HODINet are the best under all the evaluation metrics, except that $F_{\beta}$ on LFSD and $E_{\xi}$ on SSD are the second best. 
Especially on the SIP dataset, the performance gain of our method achieves 2.8\% of $F_{\beta}$, 1.6\% of $S_{\alpha}$, 1.0\% of $E_{\xi}$ and 20.4\% of ${\rm MAE}$ score, when comparing to the second best approach. 
Besides, we also improves 1.4\% of $F_{\beta}$, 1.1\% of $S_{\alpha}$, 1.4\% of $E_{\xi}$ and 24.1\% of ${\rm MAE}$ score on the DUT-RGBD dataset. As a consequence, all the results demonstrate the effectiveness of our model.

\subsubsection{Qualitative Analysis}
For qualitative comparison, we provide some representative saliency maps predicted by HODINet and other SOTA methods in Fig. \ref{fig:qualitative}. These samples include a variety of challenging scenarios, such as fine structures (1st row), fine-grained details (2nd row), large objects (3rd-4th rows), small objects (5th row), poor-quality depth maps (6th-7th rows), multiple objects (8th row), low contrast (9th row) and complex scenes (10th-11th rows). For example, we achieve structural integrity and internal consistency of the salient objects in the 2nd, 3rd and 7th images. On the contrary, other methods either miss detailed parts or mistakenly recognize the background as the foreground. 


\subsection{Ablation Study}\label{Sect:Abl}
To further investigate the effectiveness of each main component in the proposed HODINet, we conduct a series of ablation studies. These mainly include four aspects: 1) why bridging transformer-based and CNN-based networks to extract features of RGB and depth modalities; 2) how to effectively utilize the high-order attention mechanisms to discrepantly interact with different stages of feature maps; 3) whether or not can we replace the cascaded pyramid reconstruction network (CPRN) with convolutions and upsamplings to decode feature maps directly; 4) how much does the hybrid loss contribute to the performance. Note that we only modify one component at a time and retrain the networks with the same settings in Section \ref{experimental_setup}. Since the trends of experimental results found on these seven datasets are similar, we only reported ablation results on NLPR, DUT-RGBD, and LFSD in this subsection.

\begin{table*}[!t]
\centering
\caption{Ablation studies for the backbones on three datasets. $\uparrow$ ($\downarrow$) indicates the larger (smaller), the better.}
\label{tab:ablation1}
\resizebox{\linewidth}{!}{
\begin{tabular}{c|c|cccc|cccc|cccc}
\hline
 & \multirow{2}{*}{Model} & \multicolumn{4}{c|}{NLPR} & \multicolumn{4}{c|}{DUT-RGBD} & \multicolumn{4}{c}{LFSD} \\
 &  & $S_\alpha$ $\uparrow$ & MAE $\downarrow$ & $E_\xi$ $\uparrow$ & $F_\beta$ $\uparrow$ & $S_\alpha$ $\uparrow$ & MAE $\downarrow$ & $E_\xi$ $\uparrow$ & $F_\beta$ $\uparrow$ & $S_\alpha$ $\uparrow$ & MAE $\downarrow$ & $E_\xi$ $\uparrow$ & $F_\beta$ $\uparrow$ \\ \hline
 & Ours & \textbf{0.934} & \textbf{0.018} & \textbf{0.966} & \textbf{0.933} & \textbf{0.942} & \textbf{0.022} & \textbf{0.973} & \textbf{0.958} & \textbf{0.886} & \textbf{0.054} & \textbf{0.925} & \textbf{0.899} \\ \hline
\multirow{3}{*}{Backbone} & CNN+Transformer & 0.915 & 0.026 & 0.949 & 0.909 & 0.912 & 0.039 & 0.948 & 0.931 & 0.854 & 0.070 & 0.899 & 0.880 \\
 & CNN+CNN & 0.922 & 0.023 & 0.954 & 0.915 & 0.912 & 0.040 & 0.950 & 0.935 & 0.866 & 0.068 & 0.906 & 0.882 \\
 & Transformer+Transformer & 0.933 & 0.020 & 0.962 & 0.930 & 0.941 & 0.023 & 0.972 & 0.957 & 0.881 & 0.057 & 0.913 & 0.894 \\ \hline
\end{tabular}
}
\end{table*}

\begin{table*}[!t]
	\centering
	\caption{Ablation studies for the high-order fusion modules on three datasets. $\uparrow$ ($\downarrow$) indicates the larger (smaller), the better.}
	\label{tab:ablation2}
	\resizebox{\linewidth}{!}{
		\begin{tabular}{c|c|cccc|cccc|cccc}
			\hline
			& \multirow{2}{*}{Model} & \multicolumn{4}{c|}{NLPR} & \multicolumn{4}{c|}{DUT-RGBD} & \multicolumn{4}{c}{LFSD} \\
			&  & $S_\alpha$ $\uparrow$ & MAE $\downarrow$ & $E_\xi$ $\uparrow$ & $F_\beta$ $\uparrow$ & $S_\alpha$ $\uparrow$ & MAE $\downarrow$ & $E_\xi$ $\uparrow$ & $F_\beta$ $\uparrow$ & $S_\alpha$ $\uparrow$ & MAE $\downarrow$ & $E_\xi$ $\uparrow$ & $F_\beta$ $\uparrow$ \\ \hline
			& Ours & \textbf{0.934} & \textbf{0.018} & \textbf{0.966} & \textbf{0.933} & \textbf{0.942} & \textbf{0.022} & \textbf{0.973} & \textbf{0.958} & \textbf{0.886} & \textbf{0.054} & \textbf{0.925} & \textbf{0.899} \\ \hline
			\multirow{3}{*}{Fusion} & HOCF-HOSF & 0.923 & 0.025 & 0.953 & 0.922 & 0.931 & 0.030 & 0.960 & 0.948 & 0.869 & 0.064 & 0.908 & 0.879 \\
			& w/o HOCF & 0.913 & 0.035 & 0.940 & 0.909 & 0.923 & 0.032 & 0.953 & 0.940 & 0.861 & 0.062 & 0.897 & 0.873 \\
			& w/o HOSF & 0.912 & 0.034 & 0.942 & 0.910 & 0.923 & 0.037 & 0.943 & 0.938 & 0.858 & 0.071 & 0.885 & 0.873 \\ \hline
		\end{tabular}
	}
\end{table*}

\begin{table*}[!t]
	\centering
	\caption{Ablation studies for the proposed decoder and the loss function on three datasets. $\uparrow$ ($\downarrow$) indicates the larger (smaller), the better.}
	\label{tab:ablation3}
	\resizebox{\linewidth}{!}{
		\begin{tabular}{c|c|cccc|cccc|cccc}
			\hline
			& \multirow{2}{*}{Model} & \multicolumn{4}{c|}{NLPR} & \multicolumn{4}{c|}{DUT-RGBD} & \multicolumn{4}{c}{LFSD} \\
			&  & $S_\alpha$ $\uparrow$ & MAE $\downarrow$ & $E_\xi$ $\uparrow$ & $F_\beta$ $\uparrow$ & $S_\alpha$ $\uparrow$ & MAE $\downarrow$ & $E_\xi$ $\uparrow$ & $F_\beta$ $\uparrow$ & $S_\alpha$ $\uparrow$ & MAE $\downarrow$ & $E_\xi$ $\uparrow$ & $F_\beta$ $\uparrow$ \\ \hline
			& Ours & \textbf{0.934} & \textbf{0.018} & \textbf{0.966} & \textbf{0.933} & \textbf{0.942} & \textbf{0.022} & \textbf{0.973} & \textbf{0.958} & \textbf{0.886} & \textbf{0.054} & \textbf{0.925} & \textbf{0.899} \\ \hline
			Decoder & w/o CPRN & 0.921 & 0.025 & 0.952 & 0.912 & 0.927 & 0.033 & 0.956 & 0.942 & 0.872 & 0.066 & 0.903 & 0.872 \\ \hline
			Loss & only L-BCE & 0.931 & 0.023 & 0.955 & 0.927 & 0.937 & 0.029 & 0.965 & 0.952 & 0.884 & 0.059 & 0.915 & 0.893 \\ \hline
		\end{tabular}
	}
\end{table*}

\subsubsection{the necessity of using CNN-based and transformer-based networks as backbones}

\begin{figure}[!t]
	\centering
	\includegraphics[width=0.49\textwidth]{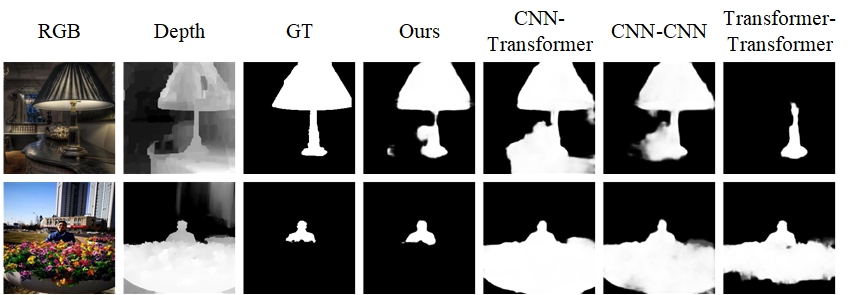}
	\centering
	\caption{Visualization results of different backbone choices.}
	\label{fig:ablation-backbone}
\end{figure}


The backbones for RGB and depth streams focus on different aspects in extracting features. In this subsection, we conduct experiments to verify the impact of different backbone choices. 
For example, if we employ CNN-based model to encode both RGB images and depth maps, the performance will decrease. 
We denote this attempt as ``CNN-CNN" in Table. \ref{tab:ablation1}. 
The reason can be explained as that PVTv2 enhances the local continuity in RGB images by overlapping patch embedding \cite{wang2021pvtv2}. 
However, compared with other CNN-based methods in Table. \ref{tab:quantitative}, we still obtain comparable results, when only using CNN-based backbones for encoding both RGB and depth streams. 
If we apply transformer-based backbone to the depth stream, the performance is slightly lower than that of CNN-based model.
Fig. \ref{fig:ablation-backbone} shows the comparison of different backbone choices.

\subsubsection{the significance of the HOSF and HOCF modules} 
\begin{figure}[!t]
	\centering
	\includegraphics[width=0.49\textwidth]{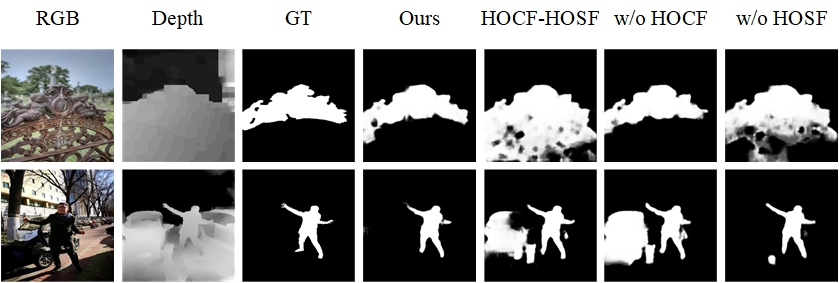}
	\centering
	\caption{Visualization of saliency detection results for different variations of fusion modules.}
	\label{fig:ablation-fusion}
\end{figure}

Complementary information between RGB images and depth maps is critical in RGB-D SOD. High-order relationship has shown its effectiveness in modeling tiny differences between images \cite{Zheng2019CVPR,Kong2017CVPR}.
In our framework, HOSF and HOCF play an important role in fusing cross-modality features. We directly substitute them with two convolutions and a concatenation operation, which are denoted as ``w/o HOSF'' and ``w/o HOCF'' in Table. \ref{tab:ablation2}. On top of that, we also exchange the sequence of HOSF and HOCF modules to investigate whether they have the same effect on different stages. Under different fusion strategies, we observe varying degrees of performance degradation in the above three datasets. This study fully indicates that the proposed high-order attention mechanisms capture critical information well in both spatial and channel dimensions. The representative examples are visualized in Fig. \ref{fig:ablation-fusion}.


\subsubsection{the effectiveness of the CPRN}
To demonstrate the importance of the proposed CPRN, we remove all cross-stage connections and directly predict the final feature maps. Specifically, we employ convolutions to reduce the number of channels to one. Then, the predictions are upsampled to the original input size. The results are shown as ``w/o CPRN'' in Table. \ref{tab:ablation3}. Compared to the complete HODINet, we find that the performance of this attempt degrades severely on all three datasets. The reason is that the top-town integration pathway decodes the adjacent feature maps gradually, preserving the consistent saliency information. In the penultimate column of Fig. \ref{fig:ablation-other}, we show some qualitative examples in this circumstance. 



\subsubsection{the usefulness of the hybrid loss} 
\begin{figure}[!t]
	\centering
	\includegraphics[width=0.49\textwidth]{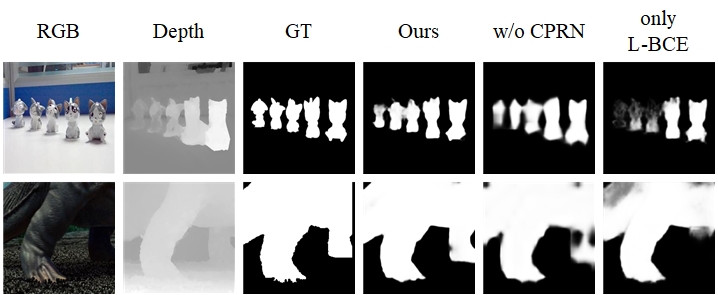}
	\centering
	\caption{Ablation study results of the proposed CPRN and the hybrid loss.}
	\label{fig:ablation-other}
\end{figure}

Several loss functions have been used in the RGB-D SOD task, achieving remarkable results \cite{DCF,9647954,CPFP}. The hybrid loss \cite{8953756} improves the integrity of salient objects with sharp and clear boundaries. In this ablation experiment, we replace the hybrid loss with the binary cross-entropy (BCE) loss. As shown in the last row ``only $\mathcal{L}^{BCE}$'' in Table. \ref{tab:ablation3}, we can see significant improvements due to the hybrid loss, especially on the $E_\xi$ metric and MAE score. 
Fig. \ref{fig:ablation-other} shows some qualitative comparisons of BCE loss and hybrid loss.

\subsection{Failure Cases and Analyses}
\begin{figure}[!t]
	\centering
	\includegraphics[width=0.4\textwidth]{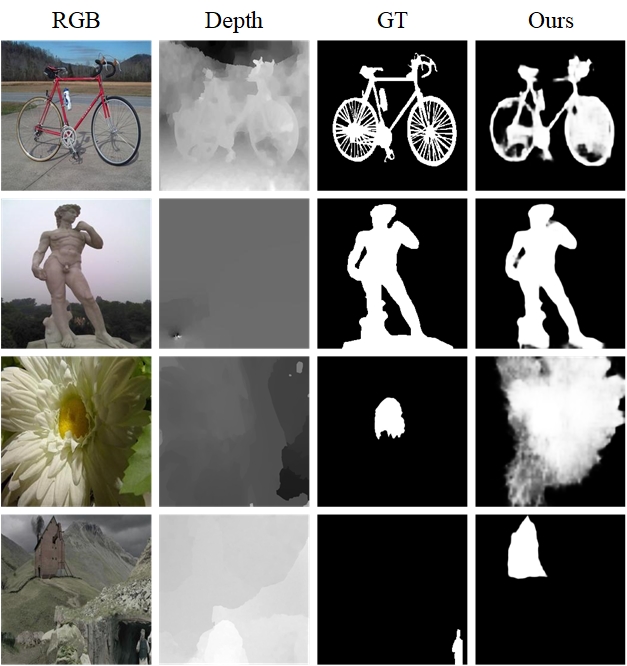}
	\centering
	\caption{Failure cases of the proposed HODINet.}
	\label{fig:fail}
\end{figure}

It is also worthwhile to analyze the reasons for failure cases. Fig. \ref{fig:fail} shows a few samples where our model fails to detect. First, the performance of the proposed method will decrease when the object is composed of fine structures. Since these contiguous pixels are easily adjacent to backgrounds, we can not obtain ideal boundaries, e.g., the first row of Fig. \ref{fig:fail}. In addition, as shown in the second and the third rows in Fig. \ref{fig:fail}, the blur depth maps provide little useful information, while the salient objects do not have explicit semantic information in the RGB images. In this circumstance, our method can not produce satisfactory results. Finally, in some scenes with multiple different objects, our model fails to detect the salient objects due to contradictory depth cues. The fourth row of Fig. \ref{fig:fail} is an example.


\section{Conclusion}
In this paper, we propose a novel high-order discrepant interaction network (HODINet) for RGB-D SOD. In the proposed framework, we adopt transformer-based and CNN-based architectures to discrepantly extract features from different modalities. Inspired by the recent success of high-order representations in other visual tasks, we design high-order spatial fusion (HOSF) and high-order channel fusion (HOCF) modules to select the most effective information. A cascaded pyramid reconstruction network (CPRN) is also employed to gradually decode the feature maps in a top-down pathway. The experimental results also prove the effectiveness of the proposed model. Compared with 24 state-of-the-art methods, HODINet achieves competitive performance on seven popular datasets.

\bibliographystyle{IEEEtran}
\bibliography{HODINet}

\vfill

\textbf{CRediT authorship contribution statement}

Kang Yi: Software, Data Curation, Writing - Original Draft. 

Jing Xu: Supervision, Resources. 

Xiao Jin: Conceptualization, Methodology, Software, Writing - Original Draft, Writing - Review \& Editing. 

Fu Guo: Validation.

Yan-Feng Wu: Validation.\\

In the final published version, Xiao Jin has renounced the authorship of this manuscript.

\vfill
\end{document}